\documentclass[a4paper,twoside]{article}

\usepackage{epsfig}
\usepackage{subcaption}
\usepackage{calc}
\usepackage{amssymb}
\usepackage{amstext}
\usepackage{amsmath}
\usepackage{amsthm}
\usepackage{multicol}
\usepackage{pslatex}
\usepackage{apalike}

\usepackage{url}

\usepackage{xcolor}
\definecolor{wong-black}        {HTML}{000000}
\definecolor{wong-lightorange}  {HTML}{E69F00}
\definecolor{wong-lightblue}    {HTML}{56B4E9}
\definecolor{wong-green}        {HTML}{009E73}
\definecolor{wong-yellow}       {HTML}{F0E442}
\definecolor{wong-darkblue}     {HTML}{0072B2}
\definecolor{wong-darkorange}   {HTML}{D55E00}
\definecolor{wong-pink}         {HTML}{CC79A7}

\usepackage{hyperref}
\hypersetup{
    colorlinks=true,
    citecolor=wong-green,
    linkcolor=wong-darkblue,
    filecolor=wong-pink,      
    urlcolor=wong-black,
    pdfpagemode=FullScreen,
    }

\usepackage{pifont}
\newcommand{\xmark}{\ding{55}}
\usepackage{array, booktabs, makecell, float}

\usepackage{SCITEPRESS}     

\newcommand\nnfootnote[1]{  
  \begin{NoHyper}
  \renewcommand\thefootnote{}\footnote{#1}%
  \addtocounter{footnote}{-1}%
  \end{NoHyper}
}

\begin{document}

\title{ad-datasets: a meta-collection of data sets for autonomous driving}

\author{\authorname{Daniel Bogdoll\sup{1}\sup{2}\textsuperscript{\textasteriskcentered}, Felix Schreyer\sup{2}\textsuperscript{\textasteriskcentered}, and J. Marius Zöllner\sup{1}\sup{2}}
\affiliation{\sup{1}FZI Research Center for Information Technology, Germany}
\affiliation{\sup{2}Karlsruhe Institute of Technology, Germany}
\email{bogdoll@fzi.de}
}


\abstract{
Autonomous driving is among the largest domains in which deep learning has been fundamental for progress within the last years. The rise of datasets went hand in hand with this development. All the more striking is the fact that researchers do not have a tool available that provides a quick, comprehensive and up-to-date overview of data sets and their features in the domain of autonomous driving.
In this paper, we present \mbox{\emph{ad-datasets}}, an online tool that provides such an overview for more than 150 data sets. The tool enables users to sort and filter the data sets according to currently 16 different categories. \mbox{\emph{ad-datasets}} is an open-source project with community contributions. It is in constant development, ensuring that the content stays up-to-date.}

\onecolumn \maketitle \normalsize

\setcounter{footnote}{0} \vfill

\section{Introduction}

\nnfootnote{\textasteriskcentered~These authors contributed equally}


One of the core building blocks on the way to fully autonomous vehicles are data sets. Of particular interest are those that contain data on all aspects of traffic. Their area of application in the research area related to autonomous driving is diverse. Hence, their number has multiplied significantly over the years. They have proven to be a necessary tool on the way to achieving the goal of autonomous vehicles. 

Given this increase in importance, it seems all the more surprising that researchers still do not have a tool at hand, that provides them an overview of existing data sets and their characteristics. 
Even today, the search for fitting data sets is a tedious and cumbersome task. Existing overviews are typically either incomplete and miss relevant data sets or come in the form of scientific papers, therefore slowly but steadily becoming outdated. This is an especially undesirable condition in such a rapidly evolving field.

Researchers are therefore regularly reliant on finding suitable data sets on their own. 
However, this task is not only extremely time-consuming, as it involves studying numerous websites and papers. It also in no way guarantees that researchers will indeed find a suitable, perhaps even optimal, data set. \\

\begin{figure*}[ht]
\centerline{\includegraphics[width=1\textwidth]{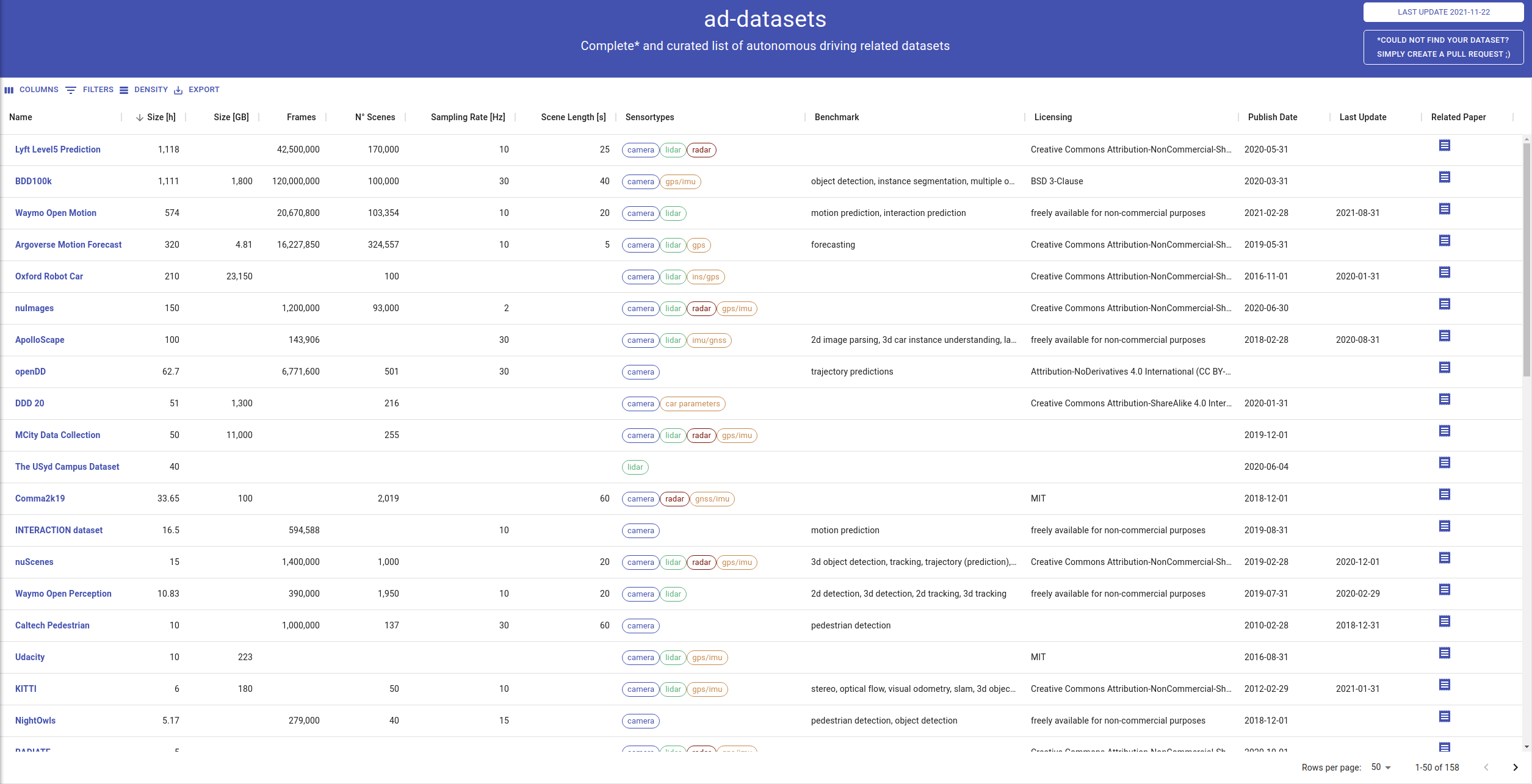}}
\caption{Screenshot of the ad-datasets web application}
\label{fig:tool}
\end{figure*}

In this publication, we present our attempt to address this unanswered challenge. With \mbox{\emph{ad-datasets}}\footnote{\url{https://ad-datasets.com/}} we have developed a tool that provides users with a comprehensive, up-to-date overview of existing data sets in the field of autonomous driving, as shown in Figure \ref{fig:tool}.
In addition, the properties of the data sets are broken down into different categories.
Users are given the opportunity to individually filter and sort the data sets according to certain criteria or rather, to their individual needs. 
Currently\footnote{\label{date}as of Nov 25, 2021}, \mbox{\emph{ad-datasets}} comprises 158 data sets.

Furthermore, \mbox{\emph{ad-datasets}} is an open-source project. The community is encouraged to contribute, such as missing data sets or metadata. We are proud to have already received several community contributions only a few weeks after the initial release, including ones from ARGO AI, a well-known company within the domain. \mbox{\emph{ad-datasets}}, hosted via Github Pages, enables this particularly easily via pull requests and automated deployments.

The further structure of this paper is organized as follows: In section \ref{sec: related work}, relevant related work is presented. Section \ref{sec: ad-datasets} introduces the tool \mbox{\emph{ad-datasets}} and its technical implementation. Section \ref{sec: evaluation} analyzes the information obtained. The final section \ref{sec: outlook} provides an outlook into future extensions.

\section{Related Work} \label{sec: related work}

\begin{table*}[t]
\centering
\begin{tabular}{lrlllrr}
\hline \hline \\
Dataset Overview & Entries & Filterable & Sortable & 
\thead{Community\\Contribution} & \thead{Number of\\Itemized\\ Properties} & Last Update      \\ \\ \hline                                                                                                  \\
ad-datasets & \textbf{158} & \textbf{Yes} & \textbf{Yes} & \textbf{Yes} & \textbf{16} & \textbf{2021} \\ \\
Bifrost     & 50 & \textbf{Yes} & \textbf{Yes} & No & - & 2020                                    \\ 
Dataset list & 25 & \textbf{Yes} & No & \textbf{Yes} & 4 & \textbf{2021}                                \\ 
Kaggle      & 31 & \textbf{Yes} & \textbf{Yes} & \textbf{Yes} & - & \textbf{2021}                       \\ 
RList       & 9 & \textbf{Yes} & \textbf{Yes} & No & 5 & \textbf{2021}                                  \\ 
Scale       & 50 & \textbf{Yes} & No & No & 4 & 2019                                              \\ 
YonoStore   & 10 & No & \textbf{Yes} & No & - & \textbf{2021}                                           \\ \\
\hline \hline \\
\end{tabular}
\caption{Comparison between \mbox{\emph{ad-datasets}} and other online tools. We compare the number of entries in each tool, the features the tools provide, the number of properties broken down in detail and when the tools were last updated.}
\label{tab:tool-comparison}
\end{table*}

The research area around autonomous driving shines with great progress and a high rate of development. Numerous advances are made every year and the amount of literature continues to grow. This is also the case in the area of data collection, respectively the area of data sets in autonomous driving. Over the years, however, the sheer amount of data sets has become increasingly complicated. For this reason, there have been attempts in the past to provide a structure to these advances. In general, these attempts can be divided into two separate categories. \\

\subsection{Scientific Papers}

First, there are studies that aim to create a general, comprehensive overview of existing data sets from the area of research. These include \cite{8317828}, where 27 data sets prevalent at the time were presented. Around three times that amount was summarized in \cite{laflamme2019driving}.

In addition to these studies, which aim directly at creating an overview of data sets, there are also studies that focus primarily on different research questions, but also contain such an overview. \cite{feng2020deep} presented their work, which also comes in an online version briefly mentioned in the following subsection, that investigated the research question of \textit{Deep Multi-modal Object Detection and Semantic Segmentation for Autonomous Driving}.
Yet, the authors also examined multi-modal data sets, with the multimodality referring to the sensors used. Compared to the works of Yin~and~Berger and Laflamme~et~al. however, this collection is much smaller in scope. \cite{heidecker2021application} takes on the topic of corner cases in highly automated driving. Here, too, a section revolves solely around suitable data sets for corner case detectors.
Finally, in the same year \cite{kim2021survey} was published that contains a survey addressing data sets for monocular 3D detection. Yet, these papers also lag behind in scope. It is important to note, that publications which do not focus on data sets but provide overviews of them as side effect typically focus strongly on their area of research. Thus, they rarely provide a broad picture of data sets and tend to focus either on popular and well known data sets or their specific niche.\\

What all these publications share is that they appeared in the format of scientific works. Thus, they have some weaknesses in common when it comes to searching for data sets. To start with, they share the problem that they become out-of-date relatively quickly. This is an undesirable characteristic, especially in a research area that is developing as quickly as the one of autonomous driving.
In addition, these overviews lack a convenient format. Naturally, the publications offer neither a filtering nor a sorting function, therefore not being as effortless and time saving as one would desire. 
 
\subsection{Online Sources}

Further, there exist various sources in online formats. These can in turn be broken down into read-only textual sources, mostly in the form of blogs \cite{choudhury-2020}\cite{spark-2018}\cite{nguyens-2021}, wiki entries \cite{unknown-author-2021}\cite{unknown-author-2020}, git repositories \cite{heyuan-2019}\cite{diaz-2021} or miscellaneous entries \cite{unknown-author-2021B}\cite{krunal-2018}\cite{feng-et-al-2021}, and into interactive tools.

The textual sources are usually kept very compact and often contain fewer than ten data sets, which are typically among the better known ones. Additionally, many times the summaries are primarily aimed at machine learning related data sets in general. It is then left to the users to filter out the relevant autonomous driving related data sets. Since all of these sources, much like scientific papers, do neither allow any filtering nor sorting functions and are typically not kept up-to-date, they are poorly suited as sources for extensive data set searches. \\

A completely different picture emerges when looking at online tools. Their format is much more suitable for searches of any kind. 

\emph{RList} \cite{unknown-author-2021C} provides such a tool. Both sorting and search functions are available to users. Further, entries are broken down into categories by release date, organization, frames and location. However, with only nine data set entries it remains rather small.  
\emph{Scale} \cite{scale-2019} provides another, much more extensive tool. It comprises 50 data sets from the areas of autonomous driving and natural language processing. Further, the tool highlights a variety of categories, namely sensor types, annotations, diversity and recording location. Filter functions are also available to the user. Unfortunately, however, the tool does not contain any data set entries after 2019 and can therefore not be considered as up-to-date.
This is where the \emph{Dataset list} \cite{plesa-2021} tool stands out. Up to the time of publication of this work, the tool was regularly updated, and can therefore be regarded as up-to-date. The tool comes with four categories of which the highlighting of the licenses differs from the ones previously described. Yet, with a volume of 25 data sets from the field of autonomous driving, it is too small to be able to claim a complete overview of existing data sets. 

Also \emph{Bifrost} with 50 data sets \cite{bifrost-2020}, \emph{Kaggle} with 31 data sets \cite{unknown-author-2021D} and \emph{YonoStore} with ten data sets \cite{yonostore-2021} cannot be seen as complete, either.
Further, unlike the aforementioned online tools, they do not provide the user with an overview of data set properties at first glance. 

More general offers such as \emph{Google Dataset Search} \cite{google-2021}, \emph{DeepAI Datasets} \cite{deepai-2021} or \emph{Papers with Code Datasets} \cite{paperswithcode-2021} have the issue that no domain-specific overview is possible. Therefore, although they list numerous data sets, they are not suitable.

Finally, it should be noted that the majority of the tools examined here do not allow the user to contribute content. Only \emph{Kaggle} and \emph{Dataset list} provide the community with an opportunity to independently add missing data sets.\\

\section{Selection Process} \label{sec: ad-datasets}

\begin{table*}[t]
\resizebox{\textwidth}{!}{%
\begin{tabular}{lrrrccccrl}
\hline \hline \\
\textbf{Name}                      & \textbf{Size {[}h{]}} & \textbf{Frames} & \textbf{N° Scenes} & \multicolumn{4}{c}{\textbf{Sensortypes}}                    & \textbf{Publish Date} & \textbf{Sources} \\
\multicolumn{4}{c}{} & \textbf{camera} & \textbf{lidar} & \textbf{radar} & \textbf{other} & \multicolumn{2}{c}{}
\\ \\ \hline \\
ApolloScape$^{i}$                    & 100                   & 143,906         &                    & \checkmark & \checkmark & & \checkmark                 & 2018.03               &\cite{wang_apolloscape_2019}\cite{unknown-author-2020C}            \\
Argoverse Motion Forecasting$^{i}$       & 320                   &                 & 324,557            & \checkmark & \checkmark & & \checkmark                      & 2019.06               & \cite{chang_argoverse_2019}\cite{unknown-author-no-dateQ}    \\
Argoverse 3D Tracking$^{i}$              &                       &                 & 113                & \checkmark & \checkmark & & \checkmark                      & 2019.06               & \cite{chang_argoverse_2019}\cite{unknown-author-no-dateQ}    \\
A2D2$^{i}$                               &                       & 433,833         & 3                  & \checkmark & \checkmark & & \checkmark                  & 2020.04               & \cite{geyer_a2d2_2020}\cite{unknown-author-no-dateO}   \\
BDD100k$^{i}$                            & 1,111                 & 120,000,000     & 100,000            & \checkmark & & & \checkmark                         & 2020.04               & \cite{yu_bdd100k_2018}\cite{unknown-author-no-dateG}                 \\
Bosch Small Traffic Lights$^{j}$  &                       & 13,427          &                    & 
\checkmark & & &                                & 2017.05               & \cite{behrendt_deep_nodate}\cite{unknown-author-no-dateZ}     \\            
Caltech Pedestrian$^{i}$                 & 10                    & 1,000,000       & 137                & \checkmark & & &                                  & 2010.03               & \cite{dollar2009pedestrian}\cite{unknown-author-no-dateP}    \\
Cityscapes 3D$^{i}$                     &                       &                 &                    & \checkmark & & & \checkmark                & 2016.02               & \cite{cordts_cityscapes_2016}\cite{gahlert_cityscapes_2020}\cite{unknown-author-2020B} \\ 
Comma2k19$^{i}$                          & 33.65                 &                 & 2,019              & \checkmark & & \checkmark & \checkmark                 & 2018.12               & \cite{schafer_commute_2018}\cite{commaai-no-date}    \\
DDD 20$^{j}$                             & 51                    &                 & 216                & \checkmark & & & \checkmark                  & 2020.02               & \cite{hu_ddd20_2020}\cite{unknown-author-2020E} \\
Fishyscapes$^{i}$                        &                       &                 &                    & \checkmark & & &                                  & 2019.09               & \cite{DBLP:journals/corr/abs-1904-03215}\cite{unknown-author-no-dateL} \\
Ford Autonomous Vehicle$^{i}$     &                       &                 &                    & 
\checkmark & \checkmark & & \checkmark                  & 2020.03               &  \cite{agarwal_ford_2020}\cite{unknown-author-no-dateR}     \\
H3D$^{j}$                                & 0.77                  & 27,721          & 160                & \checkmark & \checkmark & & \checkmark                  & 2019.03               & \cite{patil_h3d_2019} \\
India Driving$^{j}$              &                       & 10,004          & 182                & 
\checkmark & & &                                  & 2018.11               & \cite{varma2019idd}\cite{unknown-author-no-dateW}                 \\
INTERACTION$^{i}$                 & 16.5                  & 594,588         &                    & 
\checkmark & & &                                  & 2019.09               & \cite{zhan_interaction_2019}\cite{unknown-author-no-dateS} \\
KAIST Multi-Spectral Day/Night$^{i}$     &                       &                 &                    & \checkmark & \checkmark & & \checkmark & 2017.12               & \cite{8293689}\cite{unknown-author-2017} \\
KAIST Urban$^{i}$                        &                       &                 & 18                 & \checkmark & \checkmark & & \checkmark                  & 2017.09               &  \cite{jeong2019complex}\cite{unknown-author-no-dateAA}                \\
KITTI$^{i}$                              & 6                     &                 & 50                 & \checkmark & \checkmark & & \checkmark                  & 2012.03               & \cite{geiger_vision_2013}\cite{unknown-author-no-dateF}           \\
KITTI-360$^{i}$                          &                       & 400,000         &                    & \checkmark & \checkmark & & \checkmark                  & 2015.11               & \cite{xie_semantic_2016}\cite{unknown-author-no-dateK}     \\
LostAndFound$^{i}$                       &                       & 21,040          & 112                & \checkmark & & &                                  & 2016.09               & \cite{pinggera_lost_2016}\cite{unknown-author-no-dateM}  \\
Lyft Level5 Perception$^{i}$             & 2.5                   &                 & 366                & \checkmark & \checkmark & &                           & 2019.07               &  \cite{houston_one_2020}\cite{unknown-author-2021F}      \\
Lyft Level5 Prediction$^{i}$          & 1,118                 & 42,500,000      & 170,000            & \checkmark & \checkmark & \checkmark &                    & 2020.06               & \cite{houston_one_2020}\cite{unknown-author-2021E}        \\
MCity Data Collection$^{i}$              & 50                    &                 & 255                & \checkmark & \checkmark & \checkmark & \checkmark           & 2019.12               & \cite{dong_mcity_2019}   \\
NightOwls$^{j}$                          & 5.17                  & 279,000         & 40                 & \checkmark & & &                                  & 2018.12               & \cite{neumann_nightowls_2018}\cite{unknown-author-no-dateT}  \\
nuImages$^{i}$                           & 150                   & 1,200,000       & 93,000             & \checkmark & \checkmark & \checkmark & \checkmark           & 2020.07               &  \cite{unknown-author-no-dateB}                 \\
nuScenes$^{i}$                           & 15                    & 1,400,000       & 1,000              & \checkmark & \checkmark & \checkmark & \checkmark           & 2019.03               & \cite{caesar_nuscenes_2019}\cite{unknown-author-no-date}        \\
openDD$^{i}$                             & 62.7                  & 6,771,600       & 501                & \checkmark & & &                                  & 2020.09                      & \cite{DBLP:journals/corr/abs-2007-08463}\cite{unknown-author-no-dateH}         \\
Oxford Radar Robot Car$^{i}$             &                       &                 & 32                 & \checkmark & \checkmark & \checkmark & \checkmark           & 2020.02               & \cite{barnes_oxford_2020}\cite{oxford-robotics-institute-no-date}  \\
Oxford Robot Car$^{i}$                   & 210                   &                 & 100                & \checkmark & \checkmark & & \checkmark                  & 2016.11               & \cite{maddern_1_2017}\cite{unknown-author-2020D}                 \\
PandaSet$^{i}$                           & 0.23                  & 48,000          & 103                & \checkmark & \checkmark & & \checkmark                  & 2020.04               &   \cite{unknown-author-no-dateC}              \\
RadarScenes$^{j}$                        & 4                     &                 & 158                & \checkmark & & \checkmark & \checkmark                 & 2021.03               & \cite{schumann_radarscenes_2021}\cite{unknown-author-no-dateV} \\
RADIATE$^{j}$                            & 5                     &                 &                    & \checkmark & \checkmark & \checkmark & \checkmark           & 2020.10               & \cite{sheeny_radiate_2020}\cite{unknown-author-no-dateY}  \\
RoadAnomaly21$^{i}$                      &                       & 100               & 100                & \checkmark & & &                                  & 2021.04               & \cite{chan_segmentmeifyoucan_2021}\cite{unknown-author-no-dateJ}    \\
Semantic KITTI$^{i}$                     &                       & 43,552          & 21                 & 
& \checkmark & &                                   & 2019.07               & \cite{behley_semantickitti_2019}\cite{unknown-author-no-dateN}     \\
Synscapes$^{j}$                          &                       & 25,000          & 25,000             & \checkmark & & &                                  & 2018.10               & \cite{wrenninge_synscapes_2018}\cite{unknown-author-no-dateX}    \\
Udacity$^{i}$                            & 10                    &                 &                    & \checkmark & \checkmark & & \checkmark                  & 2016.09               &            \cite{udacity-no-date}      \\
Waymo Open Motion$^{i}$                  & 574                   & 20,670,800      & 103,354            & \checkmark & \checkmark & &                           & 2021.03               &  \cite{ettinger_large_2021}\cite{unknown-author-no-dateD}       \\
Waymo Open Perception$^{i}$              & 10.83                 & 390,000         & 1,950              & \checkmark & \checkmark & &                            & 2019.08               & \cite{sun_scalability_2019}\cite{unknown-author-no-dateE}        \\
WildDash$^{i}$                           &                       &                 & 156                & \checkmark & & &                                  & 2018.02               & \cite{zendel_wilddash_2018}\cite{unknown-author-no-dateI}    \\
4Seasons$^{j}$                           &                       &                 & 30                 & \checkmark & & & \checkmark                    & 2020.10               & \cite{wenzel_4seasons_2020}\cite{unknown-author-no-dateU} \\ \\ \hline\hline \\
\end{tabular}
}
\caption{Overview of the data sets analyzed in detail. Data sets marked with $^{j}$ belong to those which were chosen randomly, data sets marked with $^{i}$ have been selected deterministically by the authors}
\label{tab:data-sets}
\end{table*}

\mbox{\emph{ad-datasets}} is an online tool designed as the central point of contact for data sets in the field of autonomous driving. It includes a detailed representation of the data sets according to 16 different property categories and enables users to interact via filter and sorting functions. As of writing, \mbox{\emph{ad-datasets}} comprises 158 data sets, 40 of which were examined in detail according to the 16 categories.

\subsection{Content}

The search for data sets poses a major challenge. In this work, large and well-known data sets were easily found, both via the numerous online sources and the numerous papers which refer to them. Less known, older data sets were collected through an extensive literature research. An approach that has proven to be well-suited for their finding has been the snowballing principle.
However, newer, lesser-known publications cannot be found this way. In fact, finding them has turned out to be the greatest difficulty. Since they are rarely mentioned in literature or online sources, their search had to be designed differently. In this work, these data sets were identified through the use of search engines and via communities such as LinkedIn. It should be noted, that this procedure is associated with a high level of effort, so being well-connected in the relevant communities is of great benefit. Needless to say, this practice is not very scientific, but extremely effective, as new data sets in the community are often shared through social media.\\

In this paper, autonomous driving related data sets are defined as data sets that contain data on all aspects of traffic. They can both consist of scenes and scenarios\footnote{Following the definitions of \cite{7313256}} of road traffic or its participants. \mbox{\emph{ad-datasets}} includes, for example, data sets with video sequences of intersections from a bird's eye view, but also recordings from a vehicle directly participating in traffic. 

The primary focus of \mbox{\emph{ad-datasets}} are those data sets which were published after the famous KITTI data set, which serves, so to speak, as a time benchmark. Regarding the selection of the data sets which were analyzed in detail, the selection can be separated into two parts. 31 of the data sets were selected manually by the authors, focusing on the most popular ones. The remaining nine data sets were selected at random. The breakdown can be seen in table \ref{tab:data-sets}.

\begin{table}[h!]
\centering
\begin{tabular}{lc} \hline \hline \\
Property & Included  \\ \\ \hline \\
Annotations            & \checkmark                    \\
Benchmark              & \checkmark                    \\
Frames                 & \checkmark                    \\
Last Update            & \checkmark                    \\
Licensing              & \checkmark                    \\
Location               & \checkmark                     \\
N° Scenes              & \checkmark                    \\
Publish Date           & \checkmark                    \\
Related Datasets       & \checkmark                    \\
Related Paper          & \checkmark                    \\
Sampling Rate {[}Hz{]} & \checkmark                    \\
Scene Length {[}s{]}   & \checkmark                    \\
Sensor Types           & \checkmark                    \\
Sensors - Details      & \checkmark                    \\
Size {[}GB{]}          & \checkmark                    \\
Size {[}h{]}           & \checkmark                    \\ \\
Data Format            & \xmark                     \\
Main Focus             & \xmark                     \\ 
Recording Perspective  & \xmark                     \\
Statistics             & \xmark                     \\
Tooling                & \xmark                     \\ \\  
                                                \hline \hline \\
\end{tabular}
\caption{Property categories resulting from an expert survey with indication whether they have been included}
\label{tab:categories}
\end{table}

\subsection{Structure}

The selected property categories are in turn the result of an expert survey conducted of the research group for technical cognitive systems at the FZI Research Center for Information Technology. Over 20 different categories were suggested in the survey (Table \ref{tab:categories}). 
Ultimately, 16 categories found their way into the initial version due to time constraints. The selection of these properties that were included in the tool was made based on an examination of ten exemplary data sets. In this examination, the time required to collect the data for each property was investigated. The data was collected via the web presences of the data sets and their corresponding papers.
The decision on whether a category was included was made based upon the author's perceived importance of the property and the associated time required to include the category in the selection. The ten exemplary data sets were Cityscapes 3D, ApolloScape, Lyft Level5 Prediction, Oxford Robot Car, nuScenes, PandaSet, Waymo Open Motion, KITTI, BDD100k and openDD. The resulting 16 categories are presented in detail at this point.\\
\textbf{Annotations.}
This property describes the types of annotations with which the data sets have been provided.\\
\textbf{Benchmark.}
If benchmark challenges are explicitly listed with the data sets, they are specified here.\\
\textbf{Frames.} 
Frames states the number of frames in the data set. This includes training, test and validation data. \\
\textbf{Last Update.}
If information has been provided on updates and their dates, they can be found in this category.\\
\textbf{Licensing.}
In order to give the users an impression of the licenses of the data sets, information on them is already included in the tool.\\
\textbf{Location.}
This category lists the areas where the data sets have been recorded.  \\
\textbf{N° Scenes.} 
N° Scenes shows the number of scenes contained in the data set and includes the training, testing and validation segments. In the case of video recordings, one recording corresponds to one scene.
For data sets consisting of photos, a photo is the equivalent to a scene.\\
\textbf{Publish Date.}
The initial publication date of the data set can be found under this category. If no explicit information on the date of publication of the data set could be found, the submission date of the paper related to the set was used at this point.\\
\textbf{Related Data Sets.}
If data sets are related, the names of the related sets can be examined as well. Related data sets are, for example, those published by the same authors and building on one another.\\
\textbf{Related Paper.}
This property solely consists of a link to the paper related to the data set. \\
\textbf{Sampling Rate [Hz].} 
The Sampling Rate [Hz] property specifies the sampling rate in Hertz at which the sensors in the data set work. However, this declaration is only made if all sensors are working at the same rate or, alternatively, if the sensors are being synchronized. Otherwise, the field remains empty.\\
\textbf{Scene Length [s].}
This property describes the length of the scenes in seconds in the data set, provided all scenes have the same length. Otherwise, no information is given.
For example, if a data set has scenes with lengths between 30 and 60 seconds, no entry can be made. The background to this procedure is to maintain comparability and sortability.\\
\textbf{Sensor Types.}
This category contains a rough description of the sensor types used. Sensor types are, for example, lidar or radar.\\
\textbf{Sensors - Details.}
The Sensors - Detail category is an extension of the Sensor Types category. It includes a more detailed description of the sensors. The sensors are described in detail in terms of type and number, the frame rates they work with, the resolutions which sensors have and the horizontal field of view.\\
\textbf{Size [GB].}
The category Size [GB] describes the storage size of the data set in gigabytes.\\
\textbf{Size [h].} 
The Size [h] property is the equivalent of the Size [GB] described above, but provides information on the size of the data set in hours.
\\

It should be noted that, the name of the data set is naturally listed, too. It further acts as a link to the respective website of the data set. Further, it is worth mentioning that the general aim is to state the properties as precisely as possible. Yet, this also depends on how accurate the documentation of the data set is. For example, a size specification of 100+ hours is less precise than a specification of 103 hours. In the case of the former, the tool would only state a size of 100 hours. \\

\subsection{Technical Implementation}

\mbox{\emph{ad-datasets}} is hosted on GitHub Pages. This allows for a seamless integration of the research community. Not only can the community influence the development process, but also actively expand its content. This is especially valuable as, no matter how extensive a search, it cannot be guaranteed that the tool is indeed complete. Missing entries or metadata can be added via pull requests, which automatically trigger the build system to publish the changes.

The implementation of the tool itself was done using the frameworks React \cite{react} and Material-UI \cite{material-ui}. The latter enables a quick and uncomplicated creation of filterable as well as sortable tables.
In this work, the data grid component \cite{material-ui-datagrid} of the Material-UI framework was used.

\section{Property Examination} \label{sec: evaluation}
This section takes a closer look at the 40 data sets from \mbox{\emph{ad-datasets}} that have been analyzed in detail.

\subsection{Data sets over time}

\begin{figure*}[ht]
\centerline{\includegraphics[width=1\textwidth]{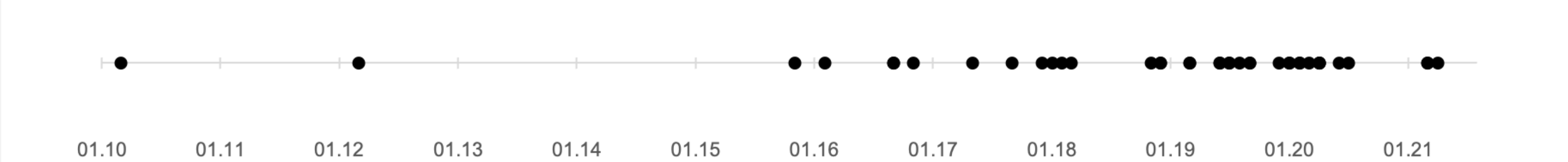}}
\caption{Timeline when data sets have been published}
\label{fig:Zeitstrahl}
\end{figure*}

In Figure \ref{fig:Zeitstrahl}, the observed data sets are shown on a timeline. While keeping in mind that the majority of the data sets were not chosen randomly, a clear tendency can be identified that both the amount of data sets and the amount of data set publications per year are steadily increasing.

For a start, this finding emphasizes once again the dynamics in the research field of autonomous driving. More to the point, it demonstrates the importance of a well-maintained and up-to-date tool in order to tackle the currently prevailing inconvenience that come with data set searches. The accelerating pace of data set publications makes it even clearer, that unsupervised overviews age at an equally increasing pace.

\subsection{Use of Sensor Types}

\begin{figure}
\centerline{\includegraphics[width=0.49\textwidth]{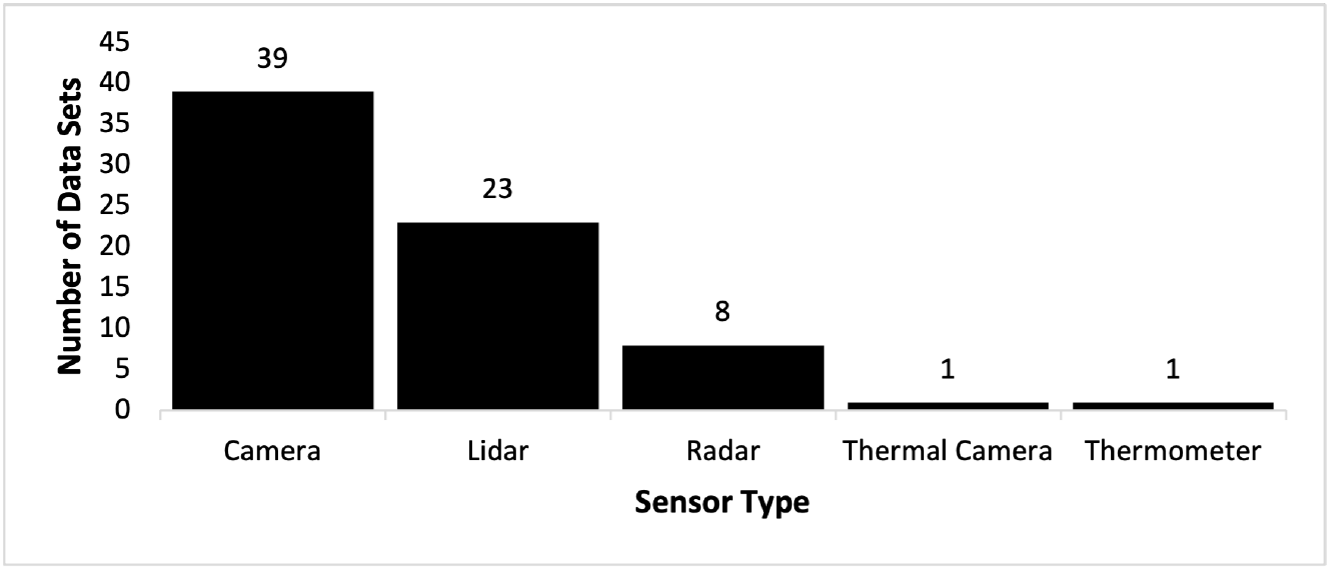}}
\caption{Frequency of use of sensor types in the data sets} 
\label{fig:Sensortypes}
\end{figure}

When looking at the sensor types used in the various data sets (Figure \ref{fig:Sensortypes}), it is first noticeable that the sensor type used most frequently is the camera. In fact, the only data set analyzed that does not utilize camera sensors is the Semantic KITTI data set. The remaining 39 out of 40 data sets make use of the camera sensor type. 

The Semantic KITTI data set in return deploys lidar sensors. 22 other data sets do the same, so that more than 50\% of the data sets include lidar data. Radar data is used much less frequently. Only eight data sets make use of the sensors. When additionally considering publication dates, it shows that radar data sets were added much later. None of the eight radar data sets have been published before the end of 2018. \newline
Yet, it seems that the importance of data sets containing radar data is increasing. Of the ten data sets published in 2020, four contained radar data.

The outsiders among the sensor types include thermal cameras and thermometers. Each appear in only one data set, the thermal camera in Cityscapes 3D and the thermometer in the Multi-Spectral Day / Night data set.

\subsection{Size}

Upon analyzing the size of data sets, one can distinguish between storage size, the time span and the number of scenes of the data sets.

\begin{figure}
\begin{minipage}{0.49\textwidth}
    \centerline{\includegraphics[width=1\textwidth]{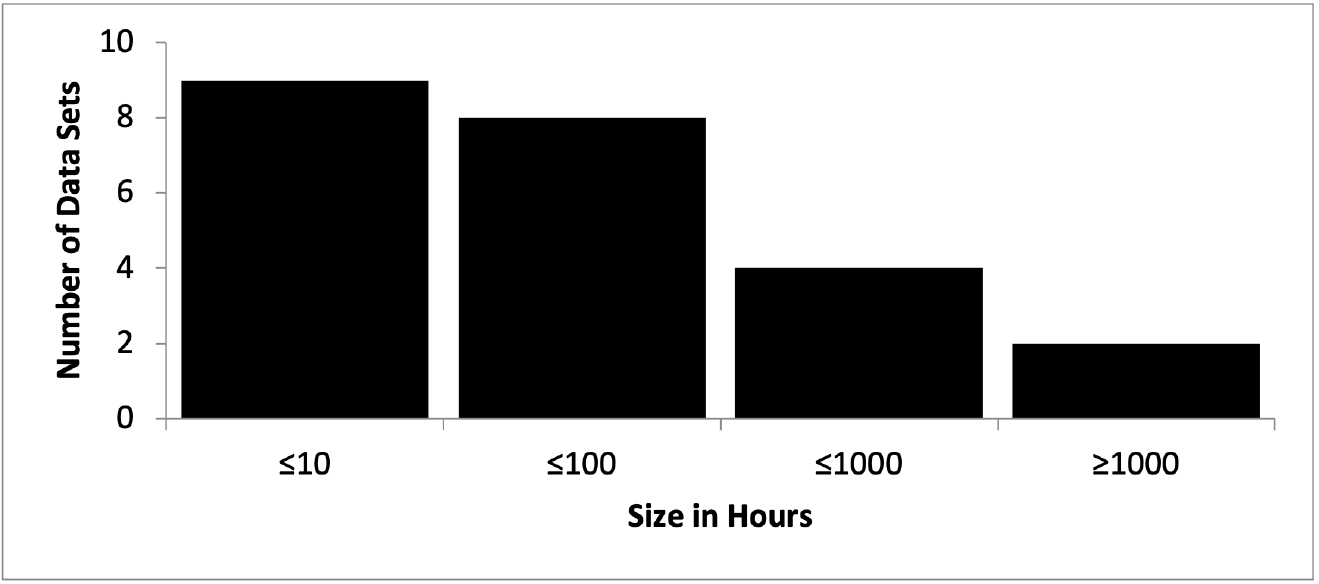}}
\end{minipage} 
\caption{Histogram depicting the distribution of the data sets over their scope of time} 
\label{fig:Histogram_h} 
\end{figure}

\begin{figure}
\centerline{\includegraphics[width=0.49\textwidth]{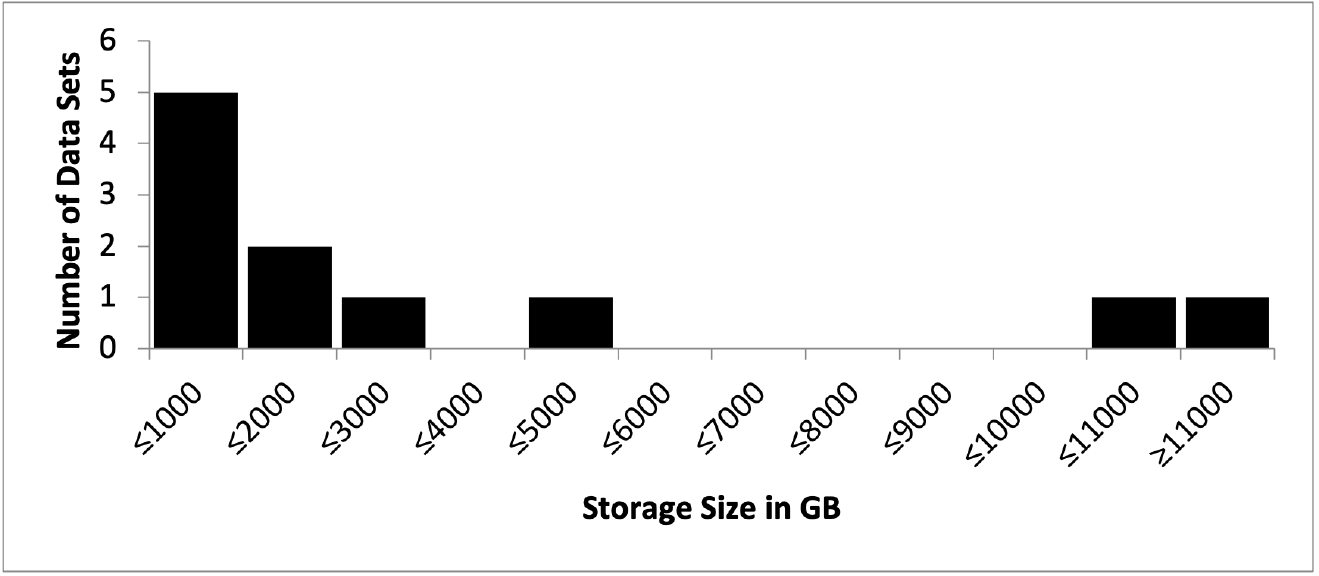}}
\caption{Histogram depicting the distribution of the data sets over their storage size} 
\label{fig:Histogram_GB}
\end{figure}

The inspection of storage size (Figure \ref{fig:Histogram_GB}) reveals that of the eleven data sets which provide information on their size, five data sets are smaller than 1,000 GB with the median being 1,300 GB. At the same time, however, there coexist significantly larger data sets. The MCity Data Collection data set is 11,000 GB large. The even larger Oxford Robot Car set is in fact with 23,150 GB more than 17 times as large as the median.
It can therefore be seen that data sets are mainly of a similar order of magnitude in terms of storage size. However, there are also sets of much larger sizes. When looking at the scope of time (Table \ref{fig:Histogram_h}), an even more pronounced picture emerges. The median of the 23 data sets, for which the corresponding information could be obtained, is 16.5 hours. There are more data sets smaller than ten hours than there are data sets larger than 100 hours. But here, too, there are examples of very large data sets. The Lyft Level5 Prediction data set is 1,118 hours large, the BDD100k data set 1,111 hours. These two data sets are over 60 times as large as the median.

On closer inspection, however, one can spot a difference between the storage wise large data sets and the time wise large data set.
Some of those data sets that are rather large in terms of storage space have been existing for multiple years. For example, the Oxford Robot Car data set was published back in 2016. Data sets which feature large scopes of time have been published rather recently. Lyft Level5 Prediction and BDD100k had not been published until 2020.

\begin{figure}
\begin{minipage}{0.49\textwidth}
\centerline{\includegraphics[width=1\textwidth]{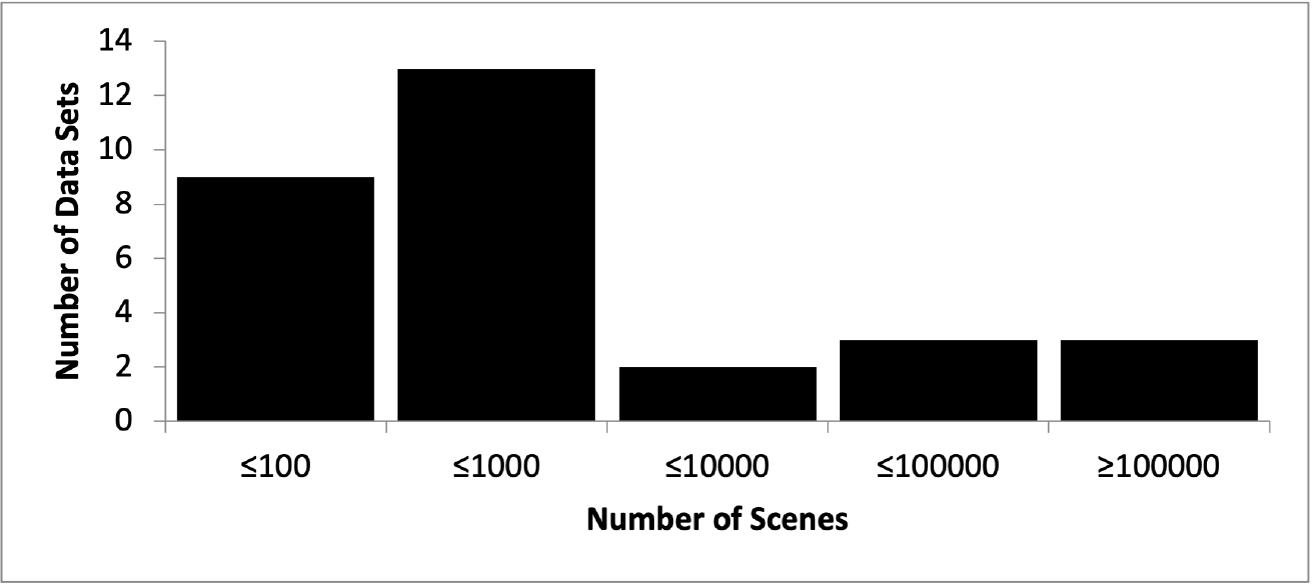}}
\end{minipage}
\caption{Histogram depicting the distribution of the data sets over their number of scenes}
\label{fig:Histogram_scenes}
\end{figure}

These impressions can also be transferred to the analysis of the number of scenes in data sets (Table \ref{fig:Histogram_scenes}). Once again, the majority revolves around a similar range regarding the number of scenes. However, again, there are outliers that are significantly larger. The median over all 30 data sets, for which information could be gathered, is 158 scenes. Opposite to that, the Lyft Level5 Prediction (170,000 scenes) and the Argoverse Motion Forecasting (324,557 scenes) are over 1,000 respectively over 2,000 times as large.

What's more, data sets that come with large number of scenes have been published relatively recently as well.


\section{Conclusion and Outlook} \label{sec: outlook}

With \mbox{\emph{ad-datasets}}, we have presented a tool that aims to simplify the previously complex and time-consuming search for suitable data sets related to the research area of autonomous driving. \mbox{\emph{ad-datasets}} offers users an overview of over 150 data sets, which are further broken down into 16 different properties. Finally, users can interact with the tool using filter and sorting functions. The timeliness of \mbox{\emph{ad-datasets}} is maintained through further maintenance of the tool and can be supported by the community, e.g., via pull requests.\\

However, the full potential of the tool has not yet been realized. To fully do so, a couple of aspects can be addressed. First, it is obviously necessary to complete the detailed analysis of the remaining data sets. This is already an ongoing work in progress. 

In addition, it is crucial to obtain feedback from the research community. After all, they are the target audience of \mbox{\emph{ad-datasets}}. Hence, their feedback is essential in developing a truly value-adding tool. Initial feedback from the community has been very positive and has already led to contributions.

Finally, it must be borne in mind that some of the properties suggested in the expert survey did not find their way into the initial version of \mbox{\emph{ad-datasets}}. Therefore, they are subject to future work. It was proposed to include information on the statistical distribution of classes, labels etc. For the time being, this proposal remains subject to future work, as there were concerns regarding copyright. Likewise, the suggested property categories data format and tooling options remain subject to future work. Important categories, which for the moment were associated with too much effort as well, are the recording perspective and the key aspects addressed by the data sets. In a later version of the tool, these will be included.

\section{Acknowledgment}
This work results partly from the KIGLIS project supported by the German Federal Ministry of Education and Research (BMBF), grant number 16KIS1231.


\bibliographystyle{apalike}
{\small
\bibliography{references}}

\begin{thebibliography}{}

\bibitem[{6D-Vision}, 2016]{unknown-author-no-dateM}
{6D-Vision} (2016).
\newblock {LostAndFoundDataset}.
\newblock \url{https://www.6d-vision.com/current-research/lostandfounddataset}.
\newblock Accessed 2022-02-03.

\bibitem[{7D Labs Inc.}, 2018]{unknown-author-no-dateX}
{7D Labs Inc.} (2018).
\newblock Synscapes.
\newblock \url{https://7dlabs.com/synscapes-overview}.
\newblock Accessed 2021-08-12.

\bibitem[{A2D2}, 2020]{unknown-author-no-dateO}
{A2D2} (2020).
\newblock Driving dataset.
\newblock \href{https://www.a2d2.audi/a2d2/en.html}.
\newblock Accessed 2021-08-22.

\bibitem[Agarwal et~al., 2020]{agarwal_ford_2020}
Agarwal, S., Vora, A., Pandey, G., Williams, W., Kourous, H., and {McBride}, J.
  (2020).
\newblock Ford multi-{AV} seasonal dataset.

\bibitem[{AIT}, 2020]{unknown-author-no-dateI}
{AIT} (2020).
\newblock Wilddash 2 benchmark.
\newblock \url{https://wilddash.cc/}.
\newblock Accessed 2021-08-09.

\bibitem[{ApolloScape}, 2018]{unknown-author-2020C}
{ApolloScape} (2018).
\newblock {ApolloScape}.
\newblock \url{http://apolloscape.auto/}.
\newblock Accessed 2021-08-04.

\bibitem[{ARGO AI}, 2019]{unknown-author-no-dateQ}
{ARGO AI} (2019).
\newblock Argoverse.
\newblock \url{https://www.argoverse.org/}.
\newblock Accessed 2021-08-22.

\bibitem[{Artisense}, 2020]{unknown-author-no-dateU}
{Artisense} (2020).
\newblock 4seasons dataset.
\newblock \url{https://www.4seasons-dataset.com/}.
\newblock Accessed 2021-08-16.

\bibitem[Autonomy and Lab, 2021]{unknown-author-no-dateAA}
Autonomy, I.~R. and Lab, P.~I. (2021).
\newblock Complex urban dataset.
\newblock \url{https://sites.google.com/view/complex-urban-dataset}.

\bibitem[Barnes et~al., 2020]{barnes_oxford_2020}
Barnes, D., Gadd, M., Murcutt, P., Newman, P., and Posner, I. (2020).
\newblock The oxford radar {RobotCar} dataset: A radar extension to the oxford
  {RobotCar} dataset.
\newblock In {\em Proceedings of the {IEEE} International Conference on
  Robotics and Automation ({ICRA})}.

\bibitem[Behley et~al., 2019]{behley_semantickitti_2019}
Behley, J., Garbade, M., Milioto, A., Quenzel, J., Behnke, S., Stachniss, C.,
  and Gall, J. (2019).
\newblock {SemanticKITTI}: A dataset for semantic scene understanding of
  {LiDAR} sequences.
\newblock In {\em Proc. of the {IEEE}/{CVF} International Conf. on Computer
  Vision ({ICCV})}.

\bibitem[Behrendt and Novak, 2017]{behrendt_deep_nodate}
Behrendt, K. and Novak, L. (2017).
\newblock A deep learning approach to traffic lights: Detection, tracking, and
  classification.
\newblock In {\em Robotics and Automation ({ICRA}), 2017 {IEEE} International
  Conference on}.

\bibitem[{Bifrost}, 2020]{bifrost-2020}
{Bifrost} (2020).
\newblock Search for visual datasets.
\newblock \url{https://datasets.bifrost.ai/}.
\newblock Accessed 2021-08-16.

\bibitem[Blum et~al., 2019]{DBLP:journals/corr/abs-1904-03215}
Blum, H., Sarlin, P., Nieto, J.~I., Siegwart, R., and Cadena, C. (2019).
\newblock The fishyscapes benchmark: Measuring blind spots in semantic
  segmentation.
\newblock {\em arXiv preprint:1904.03215}.

\bibitem[Breuer et~al., 2020]{DBLP:journals/corr/abs-2007-08463}
Breuer, A., Term{\"{o}}hlen, J., Homoceanu, S., and Fingscheidt, T. (2020).
\newblock opendd: {A} large-scale roundabout drone dataset.
\newblock {\em arXiv preprint:2007.08463}.

\bibitem[Caesar et~al., 2019]{caesar_nuscenes_2019}
Caesar, H., Bankiti, V., Lang, A.~H., Vora, S., Liong, V.~E., Xu, Q., Krishnan,
  A., Pan, Y., Baldan, G., and Beijbom, O. (2019).
\newblock {nuScenes}: A multimodal dataset for autonomous driving.
\newblock {\em {arXiv} preprint:1903.11027}.

\bibitem[{California Institute of Technology}, 2009]{unknown-author-no-dateP}
{California Institute of Technology} (2009).
\newblock {Caltech Pedestrian Detection Benchmark}.
\newblock
  \url{http://www.vision.caltech.edu/Image_Datasets/CaltechPedestrians/}.
\newblock Accessed 2021-08-22.

\bibitem[{Cambridge Spark}, 2018]{spark-2018}
{Cambridge Spark} (2018).
\newblock 50 free machine learning datasets: Self-driving cars.
\newblock
  \url{https://blog.cambridgespark.com/50-free-machine-learning-datasets-self-driving-cars-d37be5a96b28}.
\newblock {Accessed 2021-08-16}.

\bibitem[Chan et~al., 2021]{chan_segmentmeifyoucan_2021}
Chan, R., Lis, K., Uhlemeyer, S., Blum, H., Honari, S., Siegwart, R., Salzmann,
  M., Fua, P., and Rottmann, M. (2021).
\newblock {SegmentMeIfYouCan}: A benchmark for anomaly segmentation.
\newblock {\em arXiv preprint:2104.14812}.

\bibitem[Chang et~al., 2019]{chang_argoverse_2019}
Chang, M.-F., Lambert, J., Sangkloy, P., Singh, J., Bak, S., Hartnett, A.,
  Wang, D., Carr, P., Lucey, S., Ramanan, D., and Hays, J. (2019).
\newblock Argoverse: 3d tracking and forecasting with rich maps.
\newblock {\em arXiv preprint:1911.02620}.

\bibitem[Choi et~al., 2018]{8293689}
Choi, Y., Kim, N., Hwang, S., Park, K., Yoon, J.~S., An, K., and Kweon, I.~S.
  (2018).
\newblock Kaist multi-spectral day/night data set for autonomous and assisted
  driving.
\newblock {\em IEEE Transactions on Intelligent Transportation Systems}, 19(3).

\bibitem[Choudhury, 2020]{choudhury-2020}
Choudhury, A. (2020).
\newblock Top 10 popular datasets for autonomous driving projects.
\newblock
  \url{https://analyticsindiamag.com/top-10-popular-datasets-for-autonomous-driving-projects/}.
\newblock {Accessed 2021-08-18}.

\bibitem[{Cityscapes Dataset}, 2016]{unknown-author-2020B}
{Cityscapes Dataset} (2016).
\newblock Cityscapes dataset.
\newblock \url{https://www.cityscapes-dataset.com/}.
\newblock Accessed 2021-08-20.

\bibitem[{comma.ai}, 2019]{commaai-no-date}
{comma.ai} (2019).
\newblock commai/comma2k19.
\newblock \url{https://github.com/commaai/comma2k19}.
\newblock Accessed 2021-08-10.

\bibitem[Cordts et~al., 2016]{cordts_cityscapes_2016}
Cordts, M., Omran, M., Ramos, S., Rehfeld, T., Enzweiler, M., Benenson, R.,
  Franke, U., Roth, S., and Schiele, B. (2016).
\newblock The cityscapes dataset for semantic urban scene understanding.
\newblock {\em arXiv preprint:1604.01685}.

\bibitem[{cvlibs}, 2012]{unknown-author-no-dateF}
{cvlibs} (2012).
\newblock The kitti vision benchmark suite.
\newblock \url{http://www.cvlibs.net/datasets/kitti/}.
\newblock Accessed 2021-08-07.

\bibitem[{cvlibs}, 2021]{unknown-author-no-dateK}
{cvlibs} (2021).
\newblock Kitti-360: A large-scale dataset with 3d\&2d annotations.
\newblock \url{http://www.cvlibs.net/datasets/kitti-360/}.
\newblock Accessed 2021-08-21.

\bibitem[{DeepAI}, 2017]{deepai-2021}
{DeepAI} (2017).
\newblock Discover datasets for machine learning and a.i.
\newblock \url{https://deepai.org/datasets}.
\newblock Accessed 2021-08-20.

\bibitem[Diaz, 2021]{diaz-2021}
Diaz, M. (2021).
\newblock {manfreddiaz/awesome-autonomous-vehicles}.
\newblock \url{https://github.com/manfreddiaz/awesome-autonomous-vehicles}.

\bibitem[{DIY Robocars}, 2017]{unknown-author-2021B}
{DIY Robocars} (2017).
\newblock Open datasets.
\newblock \url{https://diyrobocars.com/open-datasets/}.
\newblock Accessed 2021-08-20.

\bibitem[Doll{\'a}r et~al., 2009]{dollar2009pedestrian}
Doll{\'a}r, P., Wojek, C., Schiele, B., and Perona, P. (2009).
\newblock Pedestrian detection: A benchmark.
\newblock In {\em IEEE Conference on Computer Vision and Pattern Recognition}.

\bibitem[Dong et~al., 2019]{dong_mcity_2019}
Dong, Y., Zhong, Y., Yu, W., Zhu, M., Lu, P., Fang, Y., Hong, J., and Peng, H.
  (2019).
\newblock Mcity data collection for automated vehicles study.
\newblock {\em arXiv preprint:1912.06258}.

\bibitem[{ETH VIS Group}, 2018]{unknown-author-no-dateG}
{ETH VIS Group} (2018).
\newblock Bdd100k.
\newblock \url{https://www.bdd100k.com/}.
\newblock Accessed 2021-08-07.

\bibitem[{ETH Zürich}, 2019]{unknown-author-no-dateL}
{ETH Zürich} (2019).
\newblock The fishyscapes benchmark.
\newblock \url{https://fishyscapes.com/}.
\newblock Accessed 2021-08-21.

\bibitem[Ettinger et~al., 2021]{ettinger_large_2021}
Ettinger, S., Cheng, S., Caine, B., Liu, C., Zhao, H., Pradhan, S., Chai, Y.,
  Sapp, B., Qi, C.~R., Zhou, Y., Yang, Z., Chouard, A., Sun, P., Ngiam, J.,
  Vasudevan, V., McCauley, A., Shlens, J., and Anguelov, D. (2021).
\newblock Large scale interactive motion forecasting for autonomous driving:
  The waymo open motion dataset.
\newblock In {\em Proceedings of the IEEE/CVF International Conference on
  Computer Vision (ICCV)}.

\bibitem[{Facebook}, 2013]{react}
{Facebook} (2013).
\newblock React.
\newblock \url{https://reactjs.org/}.
\newblock Accessed 2021-08-22.

\bibitem[{Facebook AI}, 2018]{paperswithcode-2021}
{Facebook AI} (2018).
\newblock Datasets.
\newblock \url{https://paperswithcode.com/datasets}.
\newblock Accessed 2021-08-16.

\bibitem[Feng et~al., 2020]{feng2020deep}
Feng, D., Haase-Sch{\"u}tz, C., Rosenbaum, L., Hertlein, H., Glaeser, C., Timm,
  F., Wiesbeck, W., and Dietmayer, K. (2020).
\newblock Deep multi-modal object detection and semantic segmentation for
  autonomous driving: Datasets, methods, and challenges.
\newblock {\em IEEE Transactions on Intelligent Transportation Systems}, 22(3).

\bibitem[Feng et~al., 2021]{feng-et-al-2021}
Feng, D., Haase-Sch{\"u}tz, C., Rosenbaum, L., Hertlein, H., Glaeser, C., Timm,
  F., Wiesbeck, W., and Dietmayer, K. (2021).
\newblock {Deep Multi-modal Object Detection and Semantic Segmentation for
  Autonomous Driving: Datasets, Methods, and Challenges}.
\newblock
  \url{https://boschresearch.github.io/multimodalperception/dataset.html}.
\newblock Accessed 2021-08-22.

\bibitem[for Image~Processing, 2017]{unknown-author-no-dateZ}
for Image~Processing, H.~C. (2017).
\newblock {Bosch Small Traffic Lights Dataset}.
\newblock
  \url{https://hci.iwr.uni-heidelberg.de/content/bosch-small-traffic-lights-dataset}.
\newblock Accessed 2021-08-12.

\bibitem[{Ford}, 2020]{unknown-author-no-dateR}
{Ford} (2020).
\newblock {Ford Autonomous Vehicle Dataset}.
\newblock \url{https://avdata.ford.com/}.
\newblock Accessed 2021-08-20.

\bibitem[{FOT-Net}, 2020]{unknown-author-2020}
{FOT-Net} (2020).
\newblock Automated driving datasets.
\newblock \url{https://wiki.fot-net.eu/index.php/Automated_Driving_Datasets}.
\newblock {Accessed 2021-08-20}.

\bibitem[Geiger et~al., 2013]{geiger_vision_2013}
Geiger, A., Lenz, P., Stiller, C., and Urtasun, R. (2013).
\newblock Vision meets robotics: The {KITTI} dataset.
\newblock {\em International Journal of Robotics Research ({IJRR})}.

\bibitem[Geyer et~al., 2020]{geyer_a2d2_2020}
Geyer, J., Kassahun, Y., Mahmudi, M., Ricou, X., Durgesh, R., Chung, A.~S.,
  Hauswald, L., Pham, V.~H., Mühlegg, M., Dorn, S., Fernandez, T., Jänicke,
  M., Mirashi, S., Savani, C., Sturm, M., Vorobiov, O., Oelker, M., Garreis,
  S., and Schuberth, P. (2020).
\newblock A2d2: Audi autonomous driving dataset.
\newblock {\em arXiv preprint:2004.06320}.

\bibitem[{Google}, 2018]{google-2021}
{Google} (2018).
\newblock Datasetsearch - autonomous driving.
\newblock
  \url{https://datasetsearch.research.google.com/search?query=Autonomous\%20driving&docid=L2cvMTFwd2Y0amZ0Yw\%3D\%3D}.
\newblock Accessed 2021-08-12.

\bibitem[Gählert et~al., 2020]{gahlert_cityscapes_2020}
Gählert, N., Jourdan, N., Cordts, M., Franke, U., and Denzler, J. (2020).
\newblock Cityscapes 3d: Dataset and benchmark for 9 {DoF} vehicle detection.
\newblock {\em arXiv preprint:2006.07864}.

\bibitem[Heidecker et~al., 2021]{heidecker2021application}
Heidecker, F., Breitenstein, J., R{\"o}sch, K., L{\"o}hdefink, J., Bieshaar,
  M., Stiller, C., Fingscheidt, T., and Sick, B. (2021).
\newblock An application-driven conceptualization of corner cases for
  perception in highly automated driving.
\newblock {\em arXiv preprin:2103.03678}.

\bibitem[{Hesai, Scale AI}, 2020]{unknown-author-no-dateC}
{Hesai, Scale AI} (2020).
\newblock {PandaSet by Hesai and Scale AI}.
\newblock \url{https://pandaset.org/}.
\newblock Accessed 2021-08-05.

\bibitem[Heyuan, 2019]{heyuan-2019}
Heyuan, L. (2019).
\newblock lhyfst/awesome-autonomous-driving-datasets.
\newblock \url{https://github.com/lhyfst/awesome-autonomous-driving-datasets}.
\newblock Accessed 2021-08-10.

\bibitem[Houston et~al., 2020]{houston_one_2020}
Houston, J., Zuidhof, G., Bergamini, L., Ye, Y., Jain, A., Omari, S.,
  Iglovikov, V., and Ondruska, P. (2020).
\newblock One thousand and one hours: Self-driving motion prediction dataset.
\newblock {\em arXiv preprint:2006.14480}.

\bibitem[Hu et~al., 2020]{hu_ddd20_2020}
Hu, Y., Binas, J., Neil, D., Liu, S.-C., and Delbrück, T. (2020).
\newblock {DDD}20 end-to-end event camera driving dataset: Fusing frames and
  events with deep learning for improved steering prediction.
\newblock {\em arXiv preprint:2005.08605}.

\bibitem[{INSAAN}, 2018]{unknown-author-no-dateW}
{INSAAN} (2018).
\newblock India driving dataset.
\newblock \url{https://idd.insaan.iiit.ac.in/}.
\newblock Accessed 2021-08-12.

\bibitem[{Inst. of Neuroinformatics, Univ. of Zurich and ETH Zurich},
  2020]{unknown-author-2020E}
{Inst. of Neuroinformatics, Univ. of Zurich and ETH Zurich} (2020).
\newblock {DDD20: end-to-end DAVIS driving dataset}.
\newblock \url{https://sites.google.com/view/davis-driving-dataset-2020/home}.

\bibitem[{INTERACTION Dataset Consortium}, 2019]{unknown-author-no-dateS}
{INTERACTION Dataset Consortium} (2019).
\newblock {INTERACTION Dataset}.
\newblock \url{https://interaction-dataset.com/}.
\newblock Accessed 2021-08-18.

\bibitem[Jeong et~al., 2019]{jeong2019complex}
Jeong, J., Cho, Y., Shin, Y.-S., Roh, H., and Kim, A. (2019).
\newblock Complex urban dataset with multi-level sensors from highly diverse
  urban environments.
\newblock {\em The International Journal of Robotics Research}, 38(6).

\bibitem[{Kaggle}, 2021]{unknown-author-2021D}
{Kaggle} (2021).
\newblock Datasets - autonomous driving.
\newblock
  \url{https://www.kaggle.com/datasets?search=Autonomous+Driving&sort=updated}.
\newblock {Accessed 2021-08-15}.

\bibitem[{KAIST}, 2017]{unknown-author-2017}
{KAIST} (2017).
\newblock {Visual Perception for Autonomous Driving}.
\newblock \url{https://sites.google.com/view/multispectral}.
\newblock Accessed 2021-08-22.

\bibitem[Kim and Hwang, 2021]{kim2021survey}
Kim, S.-h. and Hwang, Y. (2021).
\newblock A survey on deep learning based methods and datasets for monocular 3d
  object detection.
\newblock {\em Electronics}, 10(4).

\bibitem[Krunal, 2018]{krunal-2018}
Krunal (2018).
\newblock Semantic segmentation datasets for urban driving scenes.
\newblock
  \url{https://autonomous-driving.org/2018/07/15/semantic-segmentation-datasets-for-urban-driving-scenes/}.
\newblock Accessed 2021-08-22.

\bibitem[{L3Pilot}, 2019]{unknown-author-no-dateH}
{L3Pilot} (2019).
\newblock Opendd.
\newblock \url{https://l3pilot.eu/data/opendd}.
\newblock Accessed 2021-08-09.

\bibitem[Laflamme et~al., 2019]{laflamme2019driving}
Laflamme, C.-{\'E}.~N., Pomerleau, F., and Giguere, P. (2019).
\newblock Driving datasets literature review.
\newblock {\em arXiv preprint:1910.11968}.

\bibitem[{Level 5}, 2020a]{unknown-author-2021F}
{Level 5} (2020a).
\newblock Perception dataset.
\newblock \url{https://level-5.global/data/perception/}.
\newblock Accessed 2021-08-20.

\bibitem[{Level 5}, 2020b]{unknown-author-2021E}
{Level 5} (2020b).
\newblock Prediction dataset.
\newblock \url{https://level-5.global/data/prediction/}.
\newblock Accessed 2021-08-20.

\bibitem[Maddern et~al., 2017]{maddern_1_2017}
Maddern, W., Pascoe, G., Linegar, C., and Newman, P. (2017).
\newblock 1 year, 1000km: The oxford {RobotCar} dataset.
\newblock {\em The International Journal of Robotics Research ({IJRR})}, 36(1).

\bibitem[{Material-UI}, 2014]{material-ui}
{Material-UI} (2014).
\newblock {The React UI library you always wanted}.
\newblock \url{https://material-ui.com/}.
\newblock Accessed 2021-08-22.

\bibitem[{Material-UI}, 2020]{material-ui-datagrid}
{Material-UI} (2020).
\newblock Data grid.
\newblock \url{https://material-ui.com/components/data-grid/}.
\newblock Accessed 2021-08-22.

\bibitem[{Motional}, 2019a]{unknown-author-no-dateB}
{Motional} (2019a).
\newblock {nuImages}.
\newblock \url{https://www.nuscenes.org/nuimages}.
\newblock Accessed 2021-08-20.

\bibitem[{Motional}, 2019b]{unknown-author-no-date}
{Motional} (2019b).
\newblock {nuScenes}.
\newblock \url{https://www.nuscenes.org/}.
\newblock Accessed 2021-08-05.

\bibitem[Neumann et~al., 2018]{neumann_nightowls_2018}
Neumann, L., Karg, M., Zhang, S., Scharfenberger, C., Piegert, E., Mistr, S.,
  Prokofyeva, O., Thiel, R., Vedaldi, A., Zisserman, A., and Schiele, B.
  (2018).
\newblock {NightOwls}: A pedestrians at night dataset.
\newblock In {\em Asian Conference on Computer Vision}.

\bibitem[Nguyen's, 2021]{nguyens-2021}
Nguyen's, T. (2021).
\newblock List of public large-scale datasets for autonomous driving research.
\newblock
  \url{https://tin.ng/public-datasets-for-autonomous-driving-research/}.
\newblock Accessed 2021-08-22.

\bibitem[{NightOwls dataset}, 2018]{unknown-author-no-dateT}
{NightOwls dataset} (2018).
\newblock {NightOwls dataset}.
\newblock \url{https://www.nightowls-dataset.org/}.
\newblock Accessed 2021-08-18.

\bibitem[{Oxford Robotics Institute}, 2017]{unknown-author-2020D}
{Oxford Robotics Institute} (2017).
\newblock Oxford robotcar dataset.
\newblock \url{https://robotcar-dataset.robots.ox.ac.uk/}.
\newblock Accessed 2021-08-05.

\bibitem[{Oxford Robotics Institute}, 2020]{oxford-robotics-institute-no-date}
{Oxford Robotics Institute} (2020).
\newblock {Oxford Radar RobotCar Dataset}.
\newblock
  \url{https://oxford-robotics-institute.github.io/radar-robotcar-dataset/}.
\newblock Accessed 2021-08-18.

\bibitem[Patil et~al., 2019]{patil_h3d_2019}
Patil, A., Malla, S., Gang, H., and Chen, Y.-T. (2019).
\newblock The h3d dataset for full-surround 3d multi-object detection and
  tracking in crowded urban scenes.
\newblock {\em arXiv preprint:1903.01568}.

\bibitem[Pinggera et~al., 2016]{pinggera_lost_2016}
Pinggera, P., Ramos, S., Gehrig, S., Franke, U., Rother, C., and Mester, R.
  (2016).
\newblock Lost and found: Detecting small road hazards for self-driving
  vehicles.
\newblock {\em arXiv preprint:1609.04653}.

\bibitem[Plesa, 2021]{plesa-2021}
Plesa, N. (2021).
\newblock Machine learning datasets.
\newblock \url{https://www.datasetlist.com/}.
\newblock Accessed 2021-08-14.

\bibitem[{RadarScenes}, 2021]{unknown-author-no-dateV}
{RadarScenes} (2021).
\newblock {RadarScenes}.
\newblock \url{https://radar-scenes.com/}.
\newblock Accessed 2021-08-16.

\bibitem[{RList}, 2021]{unknown-author-2021C}
{RList} (2021).
\newblock List of autonomous driving open datasets.
\newblock
  \url{https://rlist.io/l/list-of-autonomous-driving-open-datasets?utm_source=insights.rlist.io&utm_medium=referral}.
\newblock Accessed 2021-08-16.

\bibitem[{Scale}, 2019]{scale-2019}
{Scale} (2019).
\newblock Open datasets.
\newblock \url{https://scale.com/open-datasets}.
\newblock Accessed 2021-08-12.

\bibitem[Schafer et~al., 2018]{schafer_commute_2018}
Schafer, H., Santana, E., Haden, A., and Biasini, R. (2018).
\newblock A commute in data: The comma2k19 dataset.

\bibitem[Schumann et~al., 2021]{schumann_radarscenes_2021}
Schumann, O., Hahn, M., Scheiner, N., Weishaupt, F., Tilly, J.~F., Dickmann,
  J., and Wöhler, C. (2021).
\newblock {RadarScenes}: A real-world radar point cloud data set for automotive
  applications.
\newblock {\em arXiv preprint:2104.02493}.

\bibitem[{Segment Me If You Can}, 2021]{unknown-author-no-dateJ}
{Segment Me If You Can} (2021).
\newblock Datasets.
\newblock \url{https://segmentmeifyoucan.com/datasets}.
\newblock Accessed 2021-08-09.

\bibitem[Sheeny et~al., 2020]{sheeny_radiate_2020}
Sheeny, M., De~Pellegrin, E., Mukherjee, S., Ahrabian, A., Wang, S., and
  Wallace, A. (2020).
\newblock {RADIATE}: A radar dataset for automotive perception.
\newblock {\em {arXiv} preprint:2010.09076}.

\bibitem[Sun et~al., 2020]{sun_scalability_2019}
Sun, P., Kretzschmar, H., Dotiwalla, X., Chouard, A., Patnaik, V., Tsui, P.,
  Guo, J., Zhou, Y., Chai, Y., Caine, B., et~al. (2020).
\newblock Scalability in perception for autonomous driving: Waymo open dataset.
\newblock In {\em Proceedings of the IEEE/CVF Conference on Computer Vision and
  Pattern Recognition}.

\bibitem[{Udacity}, 2020]{udacity-no-date}
{Udacity} (2020).
\newblock udacity/self-driving-car.
\newblock \url{https://github.com/udacity/self-driving-car/}.
\newblock Accessed 2021-08-22.

\bibitem[Ulbrich et~al., 2015]{7313256}
Ulbrich, S., Menzel, T., Reschka, A., Schuldt, F., and Maurer, M. (2015).
\newblock Defining and substantiating the terms scene, situation, and scenario
  for automated driving.
\newblock In {\em 2015 IEEE 18th International Conference on Intelligent
  Transportation Systems}.

\bibitem[{University of Bonn}, 2019]{unknown-author-no-dateN}
{University of Bonn} (2019).
\newblock {SemanticKITTI}.
\newblock \url{http://www.semantic-kitti.org/}.
\newblock Accessed 2021-08-22.

\bibitem[Varma et~al., 2019]{varma2019idd}
Varma, G., Subramanian, A., Namboodiri, A., Chandraker, M., and Jawahar, C.
  (2019).
\newblock Idd: A dataset for exploring problems of autonomous navigation in
  unconstrained environments.
\newblock In {\em IEEE Winter Conference on Applications of Computer Vision
  (WACV)}.

\bibitem[{Vision Lab, Perception and Robotics Group},
  2020]{unknown-author-no-dateY}
{Vision Lab, Perception and Robotics Group} (2020).
\newblock {Heriot-Watt RADIATE Dataset}.
\newblock \url{http://pro.hw.ac.uk/radiate/}.
\newblock Accessed 2021-08-12.

\bibitem[Wang et~al., 2019]{wang_apolloscape_2019}
Wang, P., Huang, X., Cheng, X., Zhou, D., Geng, Q., and Yang, R. (2019).
\newblock The apolloscape open dataset for autonomous driving and its
  application.
\newblock {\em {IEEE} transactions on pattern analysis and machine
  intelligence}.

\bibitem[{Waymo LLC}, 2019]{unknown-author-no-dateE}
{Waymo LLC} (2019).
\newblock {WAYMO Open Dataset - Perception}.
\newblock \url{https://waymo.com/open/data/perception/}.
\newblock Accessed 2021-08-07.

\bibitem[{Waymo LLC}, 2021]{unknown-author-no-dateD}
{Waymo LLC} (2021).
\newblock {WAYMO Open Dataset - Motion}.
\newblock \url{https://waymo.com/open/data/motion/}.
\newblock Accessed 2021-08-07.

\bibitem[Wenzel et~al., 2020]{wenzel_4seasons_2020}
Wenzel, P., Wang, R., Yang, N., Cheng, Q., Khan, Q., Stumberg, L.~v., Zeller,
  N., and Cremers, D. (2020).
\newblock 4seasons: A cross-season dataset for multi-weather {SLAM} in
  autonomous driving.
\newblock {\em arXiv preprint:2009.06364}.

\bibitem[{Wikipedia}, 2021]{unknown-author-2021}
{Wikipedia} (2021).
\newblock List of datasets for machine-learning research.
\newblock
  \url{https://en.wikipedia.org/wiki/List_of_datasets_for_machine-learning_research#Object_detection_and_recognition}.
\newblock {Accessed 2021-08-15}.

\bibitem[Wrenninge and Unger, 2018]{wrenninge_synscapes_2018}
Wrenninge, M. and Unger, J. (2018).
\newblock Synscapes: A photorealistic synthetic dataset for street scene
  parsing.
\newblock {\em arXiv preprint:1810.08705}.

\bibitem[Xie et~al., 2016]{xie_semantic_2016}
Xie, J., Kiefel, M., Sun, M.-T., and Geiger, A. (2016).
\newblock Semantic instance annotation of street scenes by 3d to 2d label
  transfer.
\newblock In {\em Conference on Computer Vision and Pattern Recognition
  ({CVPR})}.

\bibitem[Yin and Berger, 2017]{8317828}
Yin, H. and Berger, C. (2017).
\newblock When to use what data set for your self-driving car algorithm: An
  overview of publicly available driving datasets.
\newblock In {\em IEEE 20th International Conference on Intelligent
  Transportation Systems (ITSC)}.

\bibitem[{YonoStore}, 2021]{yonostore-2021}
{YonoStore} (2021).
\newblock Datasets.
\newblock
  \url{https://store.yonohub.com/product-category/datasets/?filters=product_cat[19]}.
\newblock Accessed 2021-08-12.

\bibitem[Yu et~al., 2020]{yu_bdd100k_2018}
Yu, F., Chen, H., Wang, X., Xian, W., Chen, Y., Liu, F., Madhavan, V., and
  Darrell, T. (2020).
\newblock Bdd100k: A diverse driving dataset for heterogeneous multitask
  learning.
\newblock In {\em IEEE/CVF Conference on Computer Vision and Pattern
  Recognition (CVPR)}.

\bibitem[Zendel et~al., 2018]{zendel_wilddash_2018}
Zendel, O., Honauer, K., Murschitz, M., Steininger, D., and Dominguez, G.~F.
  (2018).
\newblock {WildDash} - creating hazard-aware benchmarks.
\newblock In {\em Proceedings of the European Conference on Computer Vision
  ({ECCV})}.

\bibitem[Zhan et~al., 2019]{zhan_interaction_2019}
Zhan, W., Sun, L., Wang, D., Shi, H., Clausse, A., Naumann, M., Kümmerle, J.,
  Königshof, H., Stiller, C., Fortelle, A. d.~L., and Tomizuka, M. (2019).
\newblock {INTERACTION} dataset: An {INTERnational}, adversarial and
  cooperative {moTION} dataset in interactive driving scenarios with semantic
  maps.
\newblock {\em arXiv preprint:1910.03088}.

\end{thebibliography}

\end{document}